\documentclass[sigconf,bookmarksnumbered,unicode,colorlinks,linkcolor=blue,anchorcolor=blue,citecolor=blue]{acmart}
\usepackage{amsfonts}
\usepackage{amsmath}
\usepackage{natbib}
\usepackage{graphicx}
\usepackage{enumitem}
\usepackage{caption}
\usepackage{subcaption}
\usepackage{multirow}
\usepackage{array}
\AtBeginDocument{%
  }
\setcopyright{acmlicensed}
\copyrightyear{2025}
\acmYear{2025}
\acmDOI{XXXXXXX.XXXXXXX}
\acmConference[WWW'25 Companion]{April 28--May 02,
  2025}{Sydney, Australia}
  
\acmISBN{978-1-4503-XXXX-X/18/06}
\begin{document}
\title{LDACP: Long-Delayed Ad Conversions Prediction Model for Bidding Strategy}
\author{Peng Cui}
\authornote{Work done during an internship at Kuaishou.}
\orcid{0009-0003-0124-9043}
\email{cuipeng03@kuaishou.com}
\affiliation{%
  \institution{Beijing Institute of Technology}
  \city{Beijing}
  \country{China}
}
\author{Yiming Yang}
\email{yangyiming03@kuaishou.com}
\affiliation{%
  \institution{Kuaishou Technology}
  \city{Beijing}
  \country{China}
}
\author{Fusheng Jin}
\authornote{Corresponding author.}
\email{jfs21cn@bit.edu.cn}
\affiliation{%
  \institution{Beijing Institute of Technology}
  \city{Beijing}
  \country{China}
}
\author{Siyuan Tang}
\email{tangsiyuan@kuaishou.com}
\affiliation{%
  \institution{Kuaishou Technology}
  \city{Beijing}
  \country{China}
}
\author{Yunli Wang}
\email{wangyunli@kuaishou.com}
\affiliation{%
  \institution{Kuaishou Technology}
  \city{Beijing}
  \country{China}
}
\author{Fukang Yang}
\email{yangfukang03@kuaishou.com}
\affiliation{%
  \institution{Kuaishou Technology}
  \city{Beijing}
  \country{China}
}
\author{Yalong Jia}
\email{jiayalong@kuaishou.com}
\affiliation{%
  \institution{Kuaishou Technology}
  \city{Beijing}
  \country{China}
}
\author{Qingpeng Cai}
\email{caiqingpeng@kuaishou.com}
\affiliation{%
  \institution{Kuaishou Technology}
  \city{Beijing}
  \country{China}
}
\author{Fei Pan}
\email{panfei05@kuaishou.com}
\affiliation{%
  \institution{Kuaishou Technology}
  \city{Beijing}
  \country{China}
}
\author{Changcheng Li}
\email{lichangcheng@kuaishou.com}
\affiliation{%
  \institution{Kuaishou Technology}
  \city{Beijing}
  \country{China}
}
\author{Peng Jiang}
\email{jiangpeng@kuaishou.com}
\affiliation{%
  \institution{Kuaishou Technology}
  \city{Beijing}
  \country{China}
}
\renewcommand{\shortauthors}{Peng Cui et al.}
\begin{abstract}
  
\end{abstract}
\begin{abstract}
In online advertising, once an ad campaign is deployed, the automated bidding system dynamically adjusts the bidding strategy to optimize Cost Per Action (CPA) based on the number of ad conversions.
For ads with a long conversion delay, relying solely on the real-time tracked conversion number as a signal for bidding strategy can significantly overestimate the current CPA, leading to conservative bidding strategies. Therefore, it is crucial to predict the number of long-delayed conversions.
A typical method is to aggregate the predicted click-through conversion rate (pCTCVR) of ad impressions as the predicted ad conversion number. However, this method often results in overestimation or underestimation. 
Therefore, we propose predicting the number of conversions at the campaign level and adjusting the bidding strategy accordingly.
Nonetheless, it is challenging to predict ad conversion numbers through traditional regression methods due to the wide range of ad conversion numbers.
Previous regression works have addressed this challenge by transforming regression problems into bucket classification problems, achieving success in various scenarios. 
However, specific challenges arise when predicting the number of ad conversions:
1) The integer nature of ad conversion numbers exacerbates the discontinuity issue in one-hot hard labels;
2) The long-tail distribution of ad conversion numbers complicates tail data prediction.
In this paper, we propose the \textbf{L}ong-\textbf{D}elayed \textbf{A}d \textbf{C}onversions \textbf{P}rediction model for bidding strategy (LDACP), which consists of two sub-modules. To alleviate the issue of discontinuity in one-hot hard labels, the \textbf{B}ucket \textbf{C}lassification \textbf{M}odule with label \textbf{S}moothing method (BCMS) converts one-hot hard labels into non-normalized soft labels, then fits these soft labels by minimizing classification loss and regression loss. To address the challenge of predicting tail data, the \textbf{V}alue \textbf{R}egression \textbf{M}odule with \textbf{P}roxy labels (VRMP) uses the prediction bias of aggregated pCTCVR as proxy labels. Finally, a Mixture of Experts (MoE) structure integrates the predictions from BCMS and VRMP to obtain the final predicted ad conversion number.
Offline experiments and the online A/B test conducted on the Kuaishou platform demonstrate that our method outperforms existing competitive approaches.
\end{abstract}
\maketitle
\section{Introduction}
\begin{figure}[htbp]
    \centering
    \includegraphics[width=\linewidth]{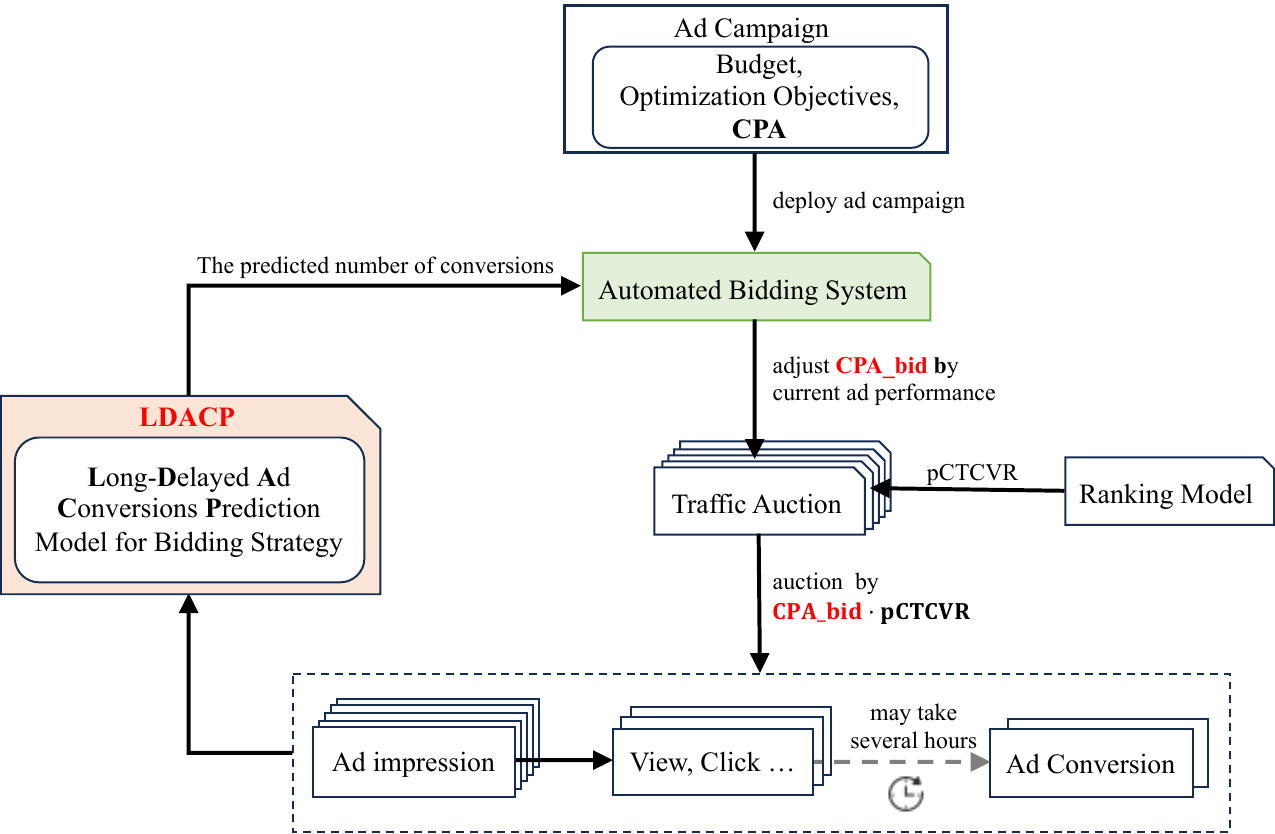}
    \caption{For ad campaigns using oCPM bidding, the automated bidding system dynamically adjusts the CPA\_bid coefficient based on the number of real-time tracked ad conversions. Our proposed LDACP predicts the number of conversions for ads with a long conversion delay. The predicted number of conversions will determine the bidding strategies of the automated bidding system.} 
    \Description{test}
    \label{fig:automatedbidding}
\end{figure}

Kuaishou is one of the top short video platforms in China, attracting hundreds of millions of daily active users. Advertisers place ads on Kuaishou to increase brand awareness and promote business growth. 
Optimized Cost Per Mille (oCPM) bidding \citep{facebook2012} is a commonly used bidding strategy.
When advertisers use oCPM bidding, they need to configure their ad campaigns \citep{zhang2023personalized}, including the budget, optimization objectives and target Cost Per Action (CPA). 
The ranking model predicts the pCTCVR (predicted click-through conversion rate) \citep{lee2018estimating}, converts advertisers' CPA bids into impression bids,  which are then used in traffic auctions.
We define the \textit{cost rate} as the ratio of actual CPA to target CPA, i.e., \textit{cost rate} = actual CPA / target CPA. To protect the interests of advertisers and ensure the profits of the platform, a \textit{cost rate} closer to 1 is preferable. On the Kuaishou platform,  the proportion of ad campaigns with a \textit{cost rate} between 0.8 and 1.2 is referred to as the \textbf{C}ompliance \textbf{R}ate (CR), which reflects the advertising cost control ability of the advertising platform.
As illustrated in Figure \ref{fig:automatedbidding}, to align the actual CPA of an ad campaign with the target CPA, the automated bidding system will adjust the campaign's CPA bid based on the real-time tracked conversion number.
However, for ads with a long conversion delay, conversions may occur days after impressions, making it impossible to track all conversions in real time. While some studies have explored how to handle the delayed feedback of conversions for conversion rate modeling \citep{chapelle2014modeling, lee2012estimating}, there is little research specifically focused on predicting the conversion number for ads with a long conversion delay.
A typical method is to aggregate the pCTCVR of ad impressions as the predicted conversion number.
This method is simple and intuitive, and when combined with pCTCVR calibration technology \cite{deng2021calibrating,wei2022posterior,mbct}, it tends to make the predicted number of conversions for the overall ad market align more closely with the actual number of conversions. 
However, this method may not guarantee the accuracy of prediction at the campaign level, potentially leading to overestimation for some ad campaigns while underestimation for others.
Furthermore, because the prediction of pCTCVR and calibration takes place before the impression, this method is unable to leverage post-impression features like the number of clicks and views, which are typically positively correlated with conversions.
To address these challenges, we propose to directly predict the number of conversions from ad impressions in a campaign, aligning it with the adjustment level of the automated bidding system, and fully leveraging post-impression features to enhance prediction accuracy.
As the number of ad conversions is an integer, we treat the prediction of ad conversion numbers at the campaign level as a regression problem.
Next, we discuss the development of regression methods and the specific challenges in our scenario.
In recent years, with the development of deep learning technology, regression methods have shifted from the traditional machine learning paradigm to the deep learning paradigm. Some literature shows that as the range of label values expands, the difficulty of regression tasks also increases. Some previous studies predict target values by fitting parameterized distributions \citep{wang2019deep, zhang2023out} or by logistic regression \citep{covington2016deep}, but these methods have strong assumptions about data distribution, and if the data distribution does not meet these assumptions, it may lead to poor performance.
The field of ordinal regression transforms regression problems into multiple binary classification problems, where each classifier predicts the probability that the label exceeds the classifier boundary.
In industry applications, particularly in the fields of Lifetime Value (LTV) estimation \citep{li2022billion}, video watch time prediction \citep{sun2024cread, lin2023tree, zhan2022deconfounding} and travel time estimation \citep{liu2023uncertainty}, some studies have attempted to transform regression problems into various forms of bucket classification problems.
\begin{figure*}[htbp]
    \centering
    \begin{subfigure}{0.49\linewidth}
        \centering
        \includegraphics[width=0.9\linewidth]{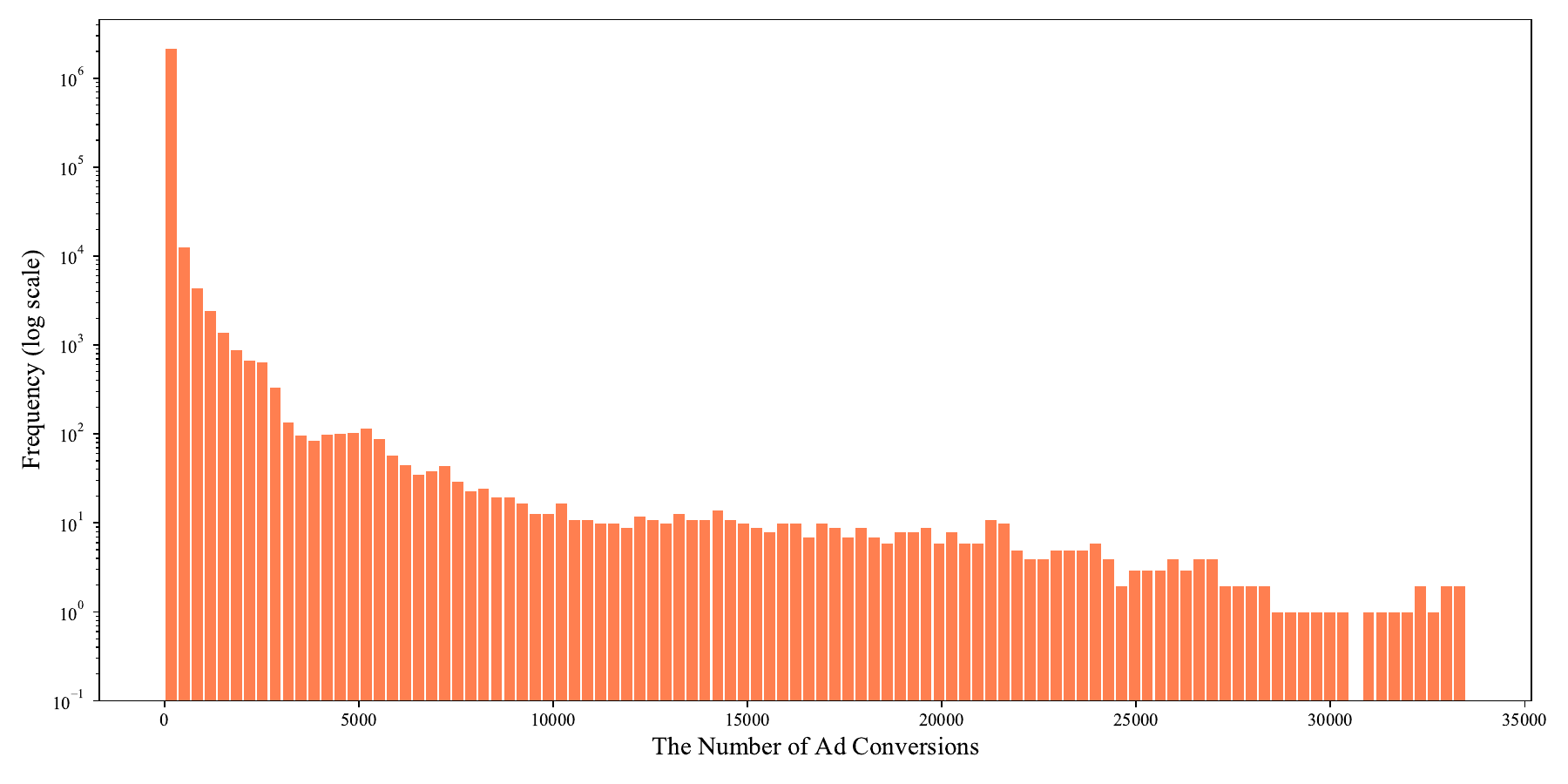}
        \caption{Ad conversion numbers exhibit a pronounced long-tail distribution (The vertical axis is on a logarithmic scale).}
        \label{fig:long_tail_dis}
    \end{subfigure}
    \hfill
    \begin{subfigure}{0.49\linewidth}
        \centering
        \includegraphics[width=0.9\linewidth]{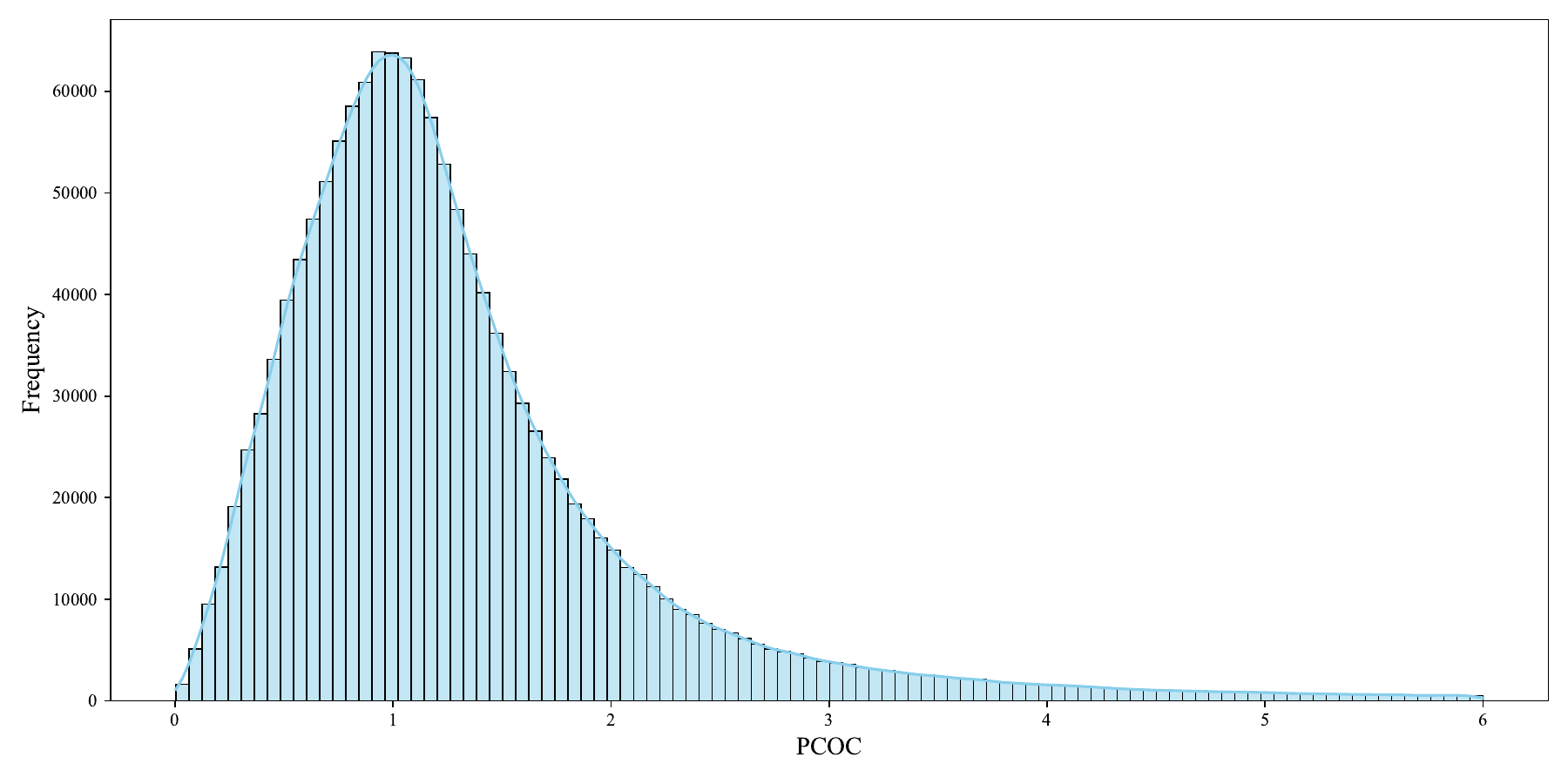}
        \caption{The PCOC of the ranking model has a narrow value range and does not exhibit a long-tail distribution.}
        \label{fig:PCOC_dis}
    \end{subfigure}
    \caption{The distribution characteristics of ad conversion numbers and the PCOC of the ranking model.}
    \label{fig:combined}
    \Description{test}
\end{figure*}
While transforming regression problems into bucket classification problems has been successful in various domains, the following challenges arise when predicting ad conversion numbers:
\begin{enumerate}[label=\arabic*.,leftmargin=*]
    \item The discontinuity issue in one-hot hard labels. Due to the integer nature of ad conversion numbers and the long-tail data distribution as illustrated in Figure \ref{fig:long_tail_dis}, under the commonly used equal-frequency bucketing condition, most label values lie near bucket boundaries. Even minor fluctuations in the label value can cause significant changes in classification labels. Therefore, the issue of discontinuity in one-hot hard labels is particularly pronounced in our scenario.
    \item Excessive width of tail buckets increases prediction difficulty for tail data. When transforming regression problems into bucket classification problems, the inference stage typically involves calculating the expectation value for each bucket. Under equal frequency bucketing, the interval span of the last bucket is very wide due to the long-tail data distribution. Utilizing the midpoint of the interval as the expectation value for the bucket may lead to a substantial overestimation of the overall data. Conversely, employing the mean of the labels within the bucket as the expectation value could result in a significant underestimation of the tail data.
\end{enumerate}
To address these challenges, we propose the \textbf{L}ong-\textbf{D}elayed \textbf{A}d \textbf{C}onversions \textbf{P}rediction model for bidding strategy (LDACP), which consists of two sub-modules: \textbf{B}ucket \textbf{C}lassification \textbf{M}odule with label \textbf{S}moothing method (BCMS) and \textbf{V}alue \textbf{R}egression \textbf{M}odule with \textbf{P}roxy labels (VRMP). The BCMS transforms one-hot hard labels into non-normalized soft labels and fits these soft labels by minimizing classification loss and regression loss, thereby addressing the issue of discontinuous one-hot hard labels. 
We use PCOC to denote the ratio of the predicted value to the actual value. The PCOC of the ranking model is the ratio of the aggregated pCTCVR of ad impressions to the actual number of ad conversions in an ad campaign, reflecting the bias of the ranking model. As illustrated in Figure \ref{fig:PCOC_dis}, the PCOC of the ranking model has a narrow value range and does not exhibit a long-tail distribution. Therefore, the VRMP employs a traditional regression method to predict the PCOC, and then converts it into the predicted ad conversion number, thus overcoming the challenge of long-tail data prediction. Finally, we use a Mixture of Experts (MoE) \citep{jacobs1991adaptive} to integrate the predictions from BCMS and VRMP, yielding the final predicted conversion number.
We conducted an offline experiment on a real-world advertising dataset collected from the Kuaishou platform. The offline experimental results show that our method outperforms existing competitive regression methods. To further verify the effectiveness of our method, we also conducted an online A/B test. The results of the online A/B test show our method improved the compliance rate by 2.29\%, increased platform revenue by 11.06\%, and boosted ad conversions by 8.68\%, bringing significant commercial value to both the advertising platform and advertisers.
Our main contributions are summarized as follows: 
\begin{enumerate}[label=\arabic*.,leftmargin=*]
    \item In online advertising, bidding strategies are influenced by the number of conversions. For ads with a long conversion delay, the number of conversions cannot be tracked in real time. We abstract this issue into the problem of predicting the number of conversions for an ad campaign. We systematically study this problem, proposing to treat it as a regression problem and leverage post-impression features to improve prediction accuracy.
    \item We propose a Long-Delayed Ad Conversions Prediction model for bidding strategy (LDACP), which consists of two sub-modules: BCMS and VRMP. The BCMS alleviates the issue of discontinuity in one-hot hard labels by converting one-hot hard labels into non-normalized soft labels and then fits those soft labels by minimizing classification loss and regression loss. The VRMP addresses the challenge of long-tail data prediction by utilizing the PCOC of the ranking model as proxy labels.
    \item Offline experiments demonstrate that our method outperforms other competitive regression methods. Furthermore, the online A/B test conducted on the Kuaishou platform confirms the effectiveness of our method in increasing the compliance rate, enhancing platform revenue, and improving ad performance.
\end{enumerate}
\section{Related Works}
Given the scarcity of literature on predicting the number of long-delayed ad conversions, our discussion of related works will primarily focus on literature concerning regression problems, which have been extensively studied across various fields.
Ordinal regression problems have been extensively researched, referring to classification tasks with ordered relationships among labels, such as age estimation, depth estimation and movie rating prediction.
Most approaches transform ordinal regression into a set of multiple binary classification tasks. 
\citep{rothe2018deep} transformed the facial age estimation problem into multiple binary classification problems, showing that converting traditional regression problems into multi-class classification problems can not only improve the accuracy but also improve the stability of training.
\citep{liu2018constrained} explicitly learned the ordinal relationships between categories using a pair-wise learning approach. 
\citep{fu2018deep} transformed continuous depth into a sequence of ordinal labels and tackled the depth estimation issue in images using ordinal regression techniques, resulting in faster convergence and higher accuracy than traditional regression approaches.
\citep{diaz2019soft} proposed converting one-hot hard labels into normalized soft labels and using Kullback-Leibler Divergence (KLD) as the loss function for training. However, the KLD loss function has certain drawbacks. For instance, when $p_{i} = 0.01$ and $q_{i} \to 0$, it can produce large gradients during back-propagation, leading to gradient explosion and severely disrupting training stability, thus it cannot be applied in our scenario. 
\citep{pan2018mean} introduced the mean-variance loss, where mean-loss penalizes the predicted difference in means, and var-loss penalizes the variance of the distribution. 
Other literature has studied regression problems in industrial applications.
\citep{wang2019deep} posited that user LTV follows a Zero-Inflated-Log-Normal (ZILN) distribution and utilized deep learning models to learn the mean and variance of the log-normal distribution. The limitation of this approach is that if the data distribution deviates from a log-normal distribution, it may be difficult to obtain optimal results. 
\citep{zhang2023out} proposed a deep learning network integrating the prediction of "Game Whale" and user LTV, using the prediction probability of "Game Whale" as gating information to guide the LTV prediction network. The modeling of LTV still involves fitting a log-normal distribution form. 
\citep{li2022billion} considered the LTV distribution to be complex and mutable, proposing a Multi Distribution Multi Experts (MDME) module based on a Divide-and-Conquer approach. 
\citep{liu2023uncertainty} estimated the distribution of travel time by transforming a single-label regression task into a multi-class classification task and devised an Adaptive Local Label Smoothing module to alleviate the over-confidence problem.
\citep{covington2016deep} utilized weighted-logical regression to predict video watch time, however, this approach necessitates that negative samples form the majority.
CREAD \citep{sun2024cread} predicted video watch time using a multiple classification approach and proposed an Error-Adaptive-Discretization module to balance classification learning loss and restoration loss.
TPM \citep{lin2023tree} advanced a single-stage classification approach to a hierarchical progressive classification approach, combining different levels of bucket classification processes for joint decision-making.
The aforementioned bucket classification methods have achieved state-of-the-art (SOTA) performance in multiple scenarios.
However, they did not explore methods to address the difficulties in predicting tail data from long-tailed distributions.
While ordinal regression research introduced a label smoothing method, the shortcomings of KLD Loss can lead to gradient explosion, impeding model training, and thus it cannot be applied to our scenario.
\section{Problem Formulation}
Assume that an ad campaign is deployed from time $t_0$. By the current time $t_k$, the campaign has $M$ impressions, and the pCTCVR of the $m$-th impression is $r_m$. Given the ad campaign profile, the number of clicks, the number of views, the number of real-time tracked conversions, etc., our goal is to predict the number of conversions resulting from impressions between time $t_0$ and time $t_k$.
We treat predicting the number of conversions of an ad campaign as a regression problem. We collect real-world ad data to build a dataset comprising $N$ samples, where each sample contains the features and label of an ad campaign from deployment to the time of bidding strategy adjustment. The features of the sample include ad campaign profile, clicks, views,  the aggregated pCTCVR $z = \sum_{m}^{M}{r_{m}}$, etc. The label of each sample is the number of conversions resulting from impressions during this period.
Let $\{x_{j}, y_{j}\}_{j=1}^{N}$ denote the dataset, where $y_{j} \in \mathcal{Y} \subset \mathbb{N}$ denotes the label of the $j$-th sample, and $x_{j} \in \mathcal{X} \subset \mathbb{R}^{d}$ denotes the input features of the $j$-th sample. Our objective is to learn a function $f$ such that the p-norm $\|f(\mathcal{X}) - \mathcal{Y}\|_{p}$ is minimized. 
\section{Method}
\subsection{Overview of LDACP}
\begin{figure}[htbp]
    \centering
    \includegraphics[width=\linewidth]{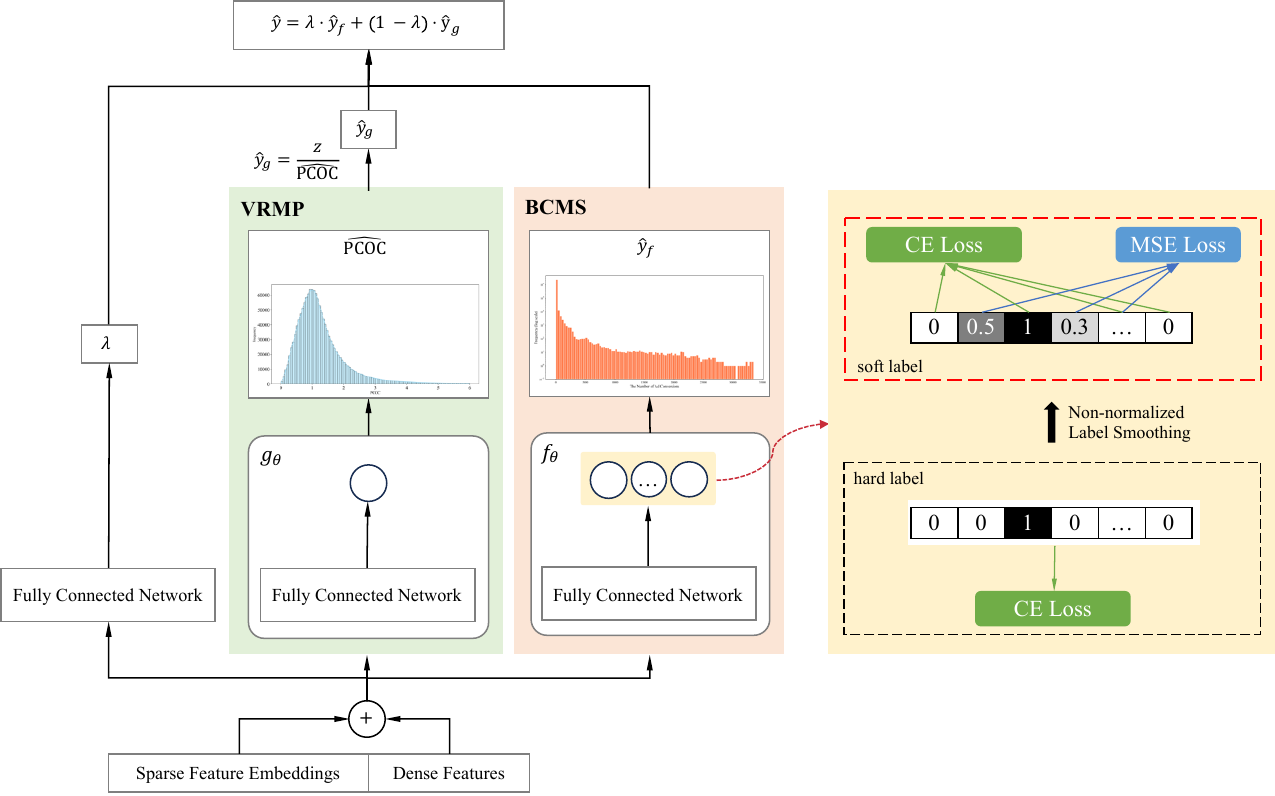}
    \caption{The LDACP consists of two sub-modules. The BCMS predicts the number of ad conversions utilizing a bucket classification method. It transforms one-hot hard labels into non-normalized soft labels, which are then fitted by minimizing Cross Entropy Loss and MSE Loss. The VRMP learns the PCOC of the ranking model using a traditional regression method, addressing the challenge of predicting tail data by leveraging the characteristic of PCOC, which does not exhibit a long-tail distribution. The predictions from BCMS and VRMP are integrated using a Mixture of Experts (MoE) structure to obtain the predicted ad conversions number $\hat{y}$.} 
    \label{fig: LDACP}
    \Description{test}
\end{figure}
As illustrated in Figure \ref{fig: LDACP}, our proposed LDACP consists of two sub-modules: BCMS and VRMP. The BCMS predicts the number of ad conversions using a bucket classification method. To address the issue of discontinuity in one-hot hard labels, BCMS converts one-hot hard labels into non-normalized soft labels and fits these soft labels by minimizing Cross Entropy Loss and MSE Loss. The output of BCMS is the predicted number of conversions, denoted as  $\hat{y}_{f}$. On the other hand, the VRMP predicts the PCOC of the ranking model via a traditional regression method and combines it with the aggregated pCTCVR $z$ to obtain the predicted ad conversion number $\hat{y}_{g}$. Subsequently, $\hat{y}_{f}$ and $\hat{y}_{g}$ are integrated through a Mixture of Experts (MoE) structure, outputting $\hat{y}$ as the final predicted number of conversions. BCMS, VRMP and MoE share the embedding parameters of sparse features.
\subsection{Bucket Classification Module with Label Smoothing Method}
\begin{table}[htb]
\caption{\textbf{Notations}}
\label{tab:notations}
\centering
\small
\begin{tabular}{cc}
\toprule
Notation & Meaning \\
\midrule
$x_{j}$ & the feature of the $j$-th sample\\
$y_{j}$ & the label of the $j$-th sample\\
$z_{j}$ &  the aggregated pCTCVR of the $j$-th sample\\
$\mathcal{T}$ & a binary tree\\
$n_{i}$  & the $i$-th node in $\mathcal{T}$\\
$e_{i \to j}$ & the edge connecting $n_{i}$ and $n_{j}$\\
$l_{i}$ & the left boundary of $n_{i}$\\
$r_{i}$ & the right boundary of $n_{i}$\\
$m_{i}$ & the boundary cut-off point of $n_{i}$\\
$\phi(y_{j})$ & the set of nodes $n_{i}$ where $l_{i} <= y_{j} < r_{i}$\\
$\mathcal{F}$ & the set of all leaf nodes in $\mathcal{T}$\\
$\mathcal{N}$ & the set of all non-leaf nodes in $\mathcal{T}$\\
\bottomrule
\end{tabular}
\end{table}
Given the success of TPM \citep{lin2023tree} in related tasks, we select TPM as the instance for BCMS and provide a detailed explanation of how to implement our label smoothing method within a binary tree structure.
Table \ref{tab:notations} provides a quick reference of all notations and their meanings used in this paper.
TPM transforms a regression problem into a multi-level bucket classification problem using a binary tree structure. 
We briefly describe the process of TPM building a binary tree.
Each node  $n_{i}$  on the tree $\mathcal{T}$ corresponds to a value range $[l_{i}, r_{i})$.       
The range $[l_{i}, r_{i})$  is divided into two sub-ranges: $[l_{i}, m_{i})$ and $[m_{i}, r_{i})$, ensuring that both sub-ranges contain an equal number of samples.
The left child $n_{2i + 1}$ corresponds to a value range of $[l_{i}, m_{i})$, and the right child $n_{2i + 2}$ corresponds to a value range of $[m_{i}, r_{i})$.
At initialization, $l_{0}$ is set to the minimum label value of the training set, and $r_{0}$ is the maximum label value.
\begin{figure}[htbp]
    \centering
    \includegraphics[width=\linewidth]{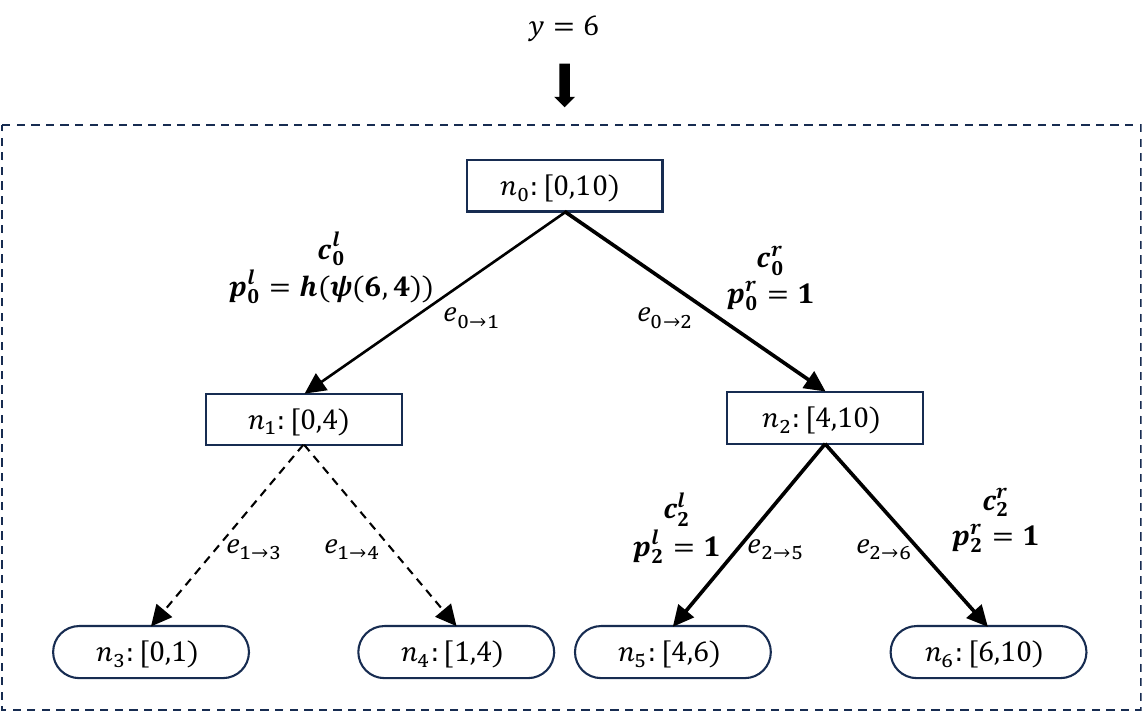}
    \caption{An instance of a TPMS. Each edge in the tree is associated with a predictor, and when $y$ changes, $h(\psi(\cdot))$ smoothly adjusts to address the discontinuity in one-hot hard labels.} 
    \label{fig:tpms}
    \Description{test}
\end{figure}
In the original TPM literature, each non-leaf node $n_{i} \in \mathcal{N}$ is associated with a binary classifier.
For a non-leaf node $n_{i}$ and a sample $(x_{j}, y_{j})$, if $y_{j} \in [l_{i}, m_{i})$, then the label of the classifier is 0;
If $y_{j} \in [m_{i}, r_{i})$, then the label of the classifier is 1. 
During back-propagation, the classifier parameters associated with  $n_{i} \in \phi(y_{j})$ are updated.
The drawback of this method is that when $y_{j} = m_{i} + \epsilon$, the label of the classifier is 1, but when $y_{j} = m_{i} - \epsilon$, the label of the classifier is 0. Here, $\epsilon$ represents a small positive number.
In other words, slight numerical fluctuations in $y$ can cause drastic changes in classification labels, disrupting the continuity of these changes.
Especially in our scenario, because ad conversion numbers are integers and exhibit a pronounced long-tail distribution, 
under the condition of equal-frequency bucketing, numerous ad conversion numbers lie near bucket boundaries. 
To address the issue of discontinuity in one-hot hard labels, we propose a novel label smoothing method. Specifically, this method converts one-hot encoded hard labels into non-normalized soft labels and then fits these soft labels by minimizing Cross Entropy Loss and MSE Loss. We implement this method in the TPM and refer to the improved model as TPMS.
\subsubsection{The Training Process of TPMS}
As illustrated in Figure \ref{fig:tpms}, let each edge on the tree be associated with a predictor. 
The predictor for the edge $e_{i \to 2i+1}$ is $c_{i}^{l}$, predicts the probability of transitioning from node $n_{i}$ to its left child $n_{2i+1}$, with the corresponding training label $p_{i}^{l}$.
The predictor for the edge $e_{i \to 2i+2}$ is $c_{i}^{r}$, predicts the probability of transitioning from node $n_{i}$ to its right child $n_{2i+2}$, with the corresponding training label $p_{i}^{r}$. \\
First, for $n_{i}$, we define the distance from $y_{j}$ to its left child $n_{2i+1}$ as:
\begin{equation} 
d_{i}^{l} = 
\begin{cases}
    0, & \text{if}\ y_{j} \in [l_{i}, m_{i}) \\
    \psi(y_{j}, m_{i}), & \text{otherwise}
\end{cases}
\end{equation}
where $\psi(y_{j}, m_{i}) \in \mathbb{R}$ is a function mapping the distance between $y_{j}$ and $m_{i}$ to a real number. \\
Second, we map this distance to a probability as follows:
\begin{equation} 
p_{i}^{l} = h(d_{i}^{l})
\end{equation}
where $h(\cdot) \in [0,1]$, and when $d_{i} < d_{j}$, $h(d_{i}) \geq h(d_{j})$.
Similarly, the distance from $y_{j}$ to the right child $n_{2i+2}$ is defined as:
\begin{equation} 
d_{i}^{r} = 
\begin{cases}
    0, & \text{if}\ y_{j} \in [m_{i}, r_{i}) \\
    \psi(y_{j}, m_{i}), & \text{otherwise}
\end{cases}
\end{equation}
This distance is also mapped to a probability:
\begin{equation} 
p_{i}^{r} = h(d_{i}^{r})
\end{equation}
Through the above steps, we convert the one-hot hard labels into non-normalized soft labels. 
When $y$ varies near $m_{i}$, $p_{i}^{l}$ and $p_{i}^{r}$ will change smoothly, thus eliminating the issue of discontinuity in one-hot hard labels near bucket boundaries. \\
Finally, we fit these soft labels by minimizing classification loss and regression loss. Let $p_{i} = \{p_{i}^{l}, p_{i}^{r}\}$. For $p_{i} \in \{0,1\}$, we treat the learning of $p_{i}$ as classification tasks, using Cross Entropy Loss as the loss function; for $p_{i} \in (0,1)$, we treat the learning of $p_{i}$ as regression tasks, using MSE Loss as the loss function, that is:
\begin{equation}
    \mathcal{L}_{BCMS} = 
    \begin{cases}
    -p_{i}\log(\hat{p}_{i}) - (1 - p_{i})\log(1 - \hat{p}_{i}), & \text{if}\ p_{i} \in \{0,1\} \\
    {(p_{i} - \hat{p_{i}})^{2}}, & \text{otherwise}
    \end{cases}
\end{equation}
\subsubsection{The Inference Process of TPMS}
First, we calculate the conditional probability from the node $n_{i}$ to its left child $n_{2i + 1}$: 
\begin{equation}
    p_{i \to 2i + 1} = \frac{\hat{p}_{i}^{l}}{\hat{p}_{i}^{l} + \hat{p}_{i}^{r}}
\end{equation}
Calculating the conditional probability from $n_{i}$ to $n_{2i + 2}$ as:
\begin{equation}
    p_{i \to 2i + 2} = 1 - p_{i \to 2i + 1}
\end{equation}
Second, we calculate the weight of the leaf node $n_{i} \in \mathcal{F}$, which is the product of the conditional probabilities from $n_{0}$ to $n_{i}$:
\begin{equation}
w_{i} = 
\begin{cases}
w_{\lfloor(i-1)/2\rfloor} \cdot p_{\lfloor (i-1)/2 \rfloor \to i}\ , & \text{if}\ i > 0 \\
1, & \text{otherwise}
\end{cases}
\end{equation}
Finally, the inference result is given by:
\begin{equation}
    \hat{y}_{f} = \sum_{n_{i} \in \mathcal{F}} e_{i} \cdot w_{i}
    \label{eq:tpm_inf}
\end{equation}
where $e_{i}$ is the expectation value of leaf node $n_{i}$, we calculate $e_{i}$ as described by \citep{fu2018deep}:
\begin{equation}
\mathcal{Y}_{i} = \{ y \in \mathcal{Y} \mid l_{i} \leq y < r_{i} \}
\label{eq:leaf_node_exp1}
\end{equation}
\begin{equation}
e_{i} = \frac{1}{|\mathcal{Y}_{i}|} \sum_{y \in \mathcal{Y}_{i}}{y}
\label{eq:leaf_node_exp2}
\end{equation}
The LDACP does not restrict the specific bucket classification method, and our label smoothing method is generally applicable to bucket classification methods with one-hot labels. The specific bucket classification method can be chosen based on the characteristics of the scenario.
\subsection{Value Regression Module with Proxy Labels}
In our scenario, ad conversion numbers exhibit a significant long-tail distribution. 
When applying the widely used equal-frequency bucketing method, the width of the last bucket tends to be very large. Using the midpoint value of the bucket range as the expectation value of the leaf node \citep{lin2023tree}, that is, $e_{i} = \frac{l_{i} + r_{i}}{2}$, can lead to a serious overestimation of the overall data. If the mean of the labels in the bucket is used as the expectation value of the leaf node \citep{rothe2018deep}, it will cause a serious underestimation of the tail data. If the interval percentage regression is used to predict the expectation value of the leaf node \citep{li2022billion}, the accuracy requirement for the percentage regression will be very high due to the wide interval range.
To overcome the challenge of predicting tail data, we use the PCOC of the ranking model as proxy labels. Since the PCOC has a narrow value range and does not exhibit a long-tail distribution, the VRMP learns the PCOC through the traditional regression method, outputting the predicted $\hat{PCOC}$. Subsequently, this $\hat{PCOC}$ is converted into the predicted number of conversions $\hat{y}_{g}$. The detailed steps is as follows: \\
First, we calculate the PCOC of the ranking model as:
\begin{equation}
PCOC = 
\begin{cases}
    \frac{z}{y}, & \text{if } y > 0  \\
    1, & \text{otherwise}
\end{cases}
\label{eq:pcoc}
\end{equation}
As illustrated in Figure \ref{fig:PCOC_dis}, the range of PCOC is narrower compared to the number of conversions, and it does not exhibit a long-tail distribution. 
Therefore, we propose using it as the learning label and employing a traditional regression method to predict it:
\begin{equation}
\hat{PCOC} = g_{\theta}(x, z)
\label{eq:pcoc_reg}
\end{equation}
Second, the MAE Loss is employed to learn the PCOC:
\begin{equation}
\mathcal{L}_{VRMP} = |\hat{PCOC} - PCOC|
\end{equation}
Finally, we convert the $\hat{PCOC} $ into the predicted ad conversion number: 
\begin{equation}
\hat{y}_{g} = \frac{z}{\hat{PCOC}}
\end{equation}
Through the above steps, we obtained the predicted number of ad conversions $\hat{y}_{g}$ by using PCOC as proxy labels.
\subsection{Integrating BCMS and VRMP via MoE}
We adopt a MoE structure, predicting the gating signal $\lambda \in [0,1]$ to integrate $\hat{y}_{f}$ and $\hat{y}_{g}$ as follows:
\begin{equation}
\hat{y} = \lambda \cdot \hat{y}_{f} + (1 - \lambda) \cdot \hat{y}_{g}
\label{eq:MoE}
\end{equation}
To prevent the model from being dominated by samples with large labels, we employ a loss function in the form of MAPE to balance the weights of all samples:
\begin{equation}
\mathcal{L}_{MoE} = | \lambda \cdot \frac{\hat{y}_{f}}{y} + (1 - \lambda) \cdot \frac{\hat{y}_{g}}{y} - 1|
\label{eq:moe_mape}
\end{equation}
The overall loss function of the LDACP is:
\begin{equation}
\mathcal{L}_{TOTAL} = \mathcal{L}_{BCMS} + \alpha \cdot \mathcal{L}_{VRMP} + \beta \cdot \mathcal{L}_{MoE}
\end{equation}
where $\alpha$ and $\beta$ are the hyperparameters that balance $\mathcal{L}_{BCMS}$, $\mathcal{L}_{VRMP}$ and $\mathcal{L}_{MoE}$.
\section{Experiment}
\label{sec:experiment}
To evaluate the effectiveness of our method, we conducted both offline and online experiments.
\subsection{Offline Experiment}
\subsubsection{Dataset}
\begin{table}[htbp]
    \caption{Statistics of Kuai-AD}
    \centering
    \begin{tabular}{lc}
        \toprule
        maximum number of conversions& 33492 \\
        minimum number of conversions& 0 \\
        average number of conversions& 24.70 \\
        number of product categories& 1141 \\
        sample size & 2738705 \\
        \bottomrule
    \end{tabular}
    \label{tab:dataset}
\end{table}
\renewcommand{\arraystretch}{1.2}
\begin{table*}[htbp]
    \caption{Performance on Kuai-AD dataset. We use the number of conversions and PCOC as training labels, denoted as method-N and method-P respectively. The second-best results are underlined, and the best results are highlighted in bold.}
    \footnotesize
    \centering
    \begin{tabular}{cccccccccccccc}
      \hline
      \textbf{Metric}                     & \textbf{Industry} & \textbf{RM}  & \textbf{VR-N}   & \textbf{VR-P}     & \textbf{ZILN-N} & \textbf{ZILN-P} & \textbf{CREAD-N}  & \textbf{CREAD-P}  & \textbf{MDME-N} & \textbf{MDME-P}  & \textbf{TPM-N}    & \textbf{TPM-P} & \textbf{LDACP}   \\  
      \hline
      \multirow{5}{*}{MAPE $\downarrow$}  & Games             & 0.9041       & 0.4101          & 0.4160            & 0.3853          & 0.9893          & 0.5032            & 0.3694            & 0.4817          &\underline{0.3464}& 0.3812            & 0.4263         & \textbf{0.2746} \\
                                          & Media             & 0.5065       & 0.3245          & 0.2817            & 0.2815          & 0.5207          & 0.4285            &\underline{0.2689} & 0.2977          & 0.4419           & 0.2778            & 0.3285         & \textbf{0.2011} \\
                                          & E-commerce        & 0.5157       & 0.3607          & 0.3445            & 0.3309          & 0.7023          & 0.3405            &\underline{0.2892} & 0.3045          & 0.3200           & 0.4333            & 0.4099         & \textbf{0.2433} \\
                                          & Life Services     & 0.5214       & 0.3308          & 0.3041            & 0.3079          & 0.5411          & 0.3463            &\underline{0.2998} & 0.3260          & 0.3519           & 0.3536            & 0.3899         & \textbf{0.2293} \\
                                          & Total             & 0.5604       & 0.3436          & 0.3129            & 0.3083          & 0.5395          & 0.4147            &\underline{0.2914} & 0.3266          & 0.3554           & 0.3262            & 0.3453         & \textbf{0.2228} \\
      \hline
      \multirow{5}{*}{CR $\uparrow$}      & Games             & 24.31\%       & 33.37\%          &\underline{42.18\%} & 38.54\%          & 31.16\%          & 28.85\%            & 40.77\%            & 40.58\%          & 41.84\%           & 39.89\%            & 35.92\%         & \textbf{53.08\%}          \\
                                          & Media             & 27.93\%       & 42.14\%          &\underline{56.79\%} & 52.77\%          & 42.58\%          & 35.41\%            & 56.22\%            & 55.99\%          & 56.43\%           & 55.09\%            & 47.99\%         & \textbf{65.64\%}          \\
                                          & E-commerce        & 33.22\%       & 45.58\%          & 49.76\%            & 52.06\%          & 40.65\%          & 47.56\%            &\underline{54.93\%} & 52.16\%          & 51.87\%           & 47.07\%            & 49.22\%         & \textbf{59.06\%}          \\
                                          & Life Services     & 33.83\%       & 48.56\%          & 55.77\%            & 56.54\%          & 38.27\%          & 49.55\%            &\underline{58.91\%} & 56.91\%          & 56.52\%           & 56.37\%            & 49.05\%         & \textbf{65.65\%}          \\
                                          & Total             & 28.88\%       & 42.12\%          & 53.21\%            & 50.85\%          & 41.44\%          & 38.07\%            &\underline{53.80\%} & 52.99\%          & 53.36\%           & 51.57\%            & 48.77\%         & \textbf{62.32\%}          \\
      \hline
    \end{tabular}
    \label{tab:offline_exp}
\end{table*}
We collected real-world advertising data from the Kuaishou platform as our offline experiment dataset (referred to as Kuai-AD). 
Kuai-AD comprises advertising data spanning 8 consecutive days. The data from the first seven days were randomly divided into training and test sets in a ratio of 9:1, while the data from the last day was used as the test set. 
The statistics of Kuai-AD are presented in Table \ref{tab:dataset}. 
To account for the delayed ad conversions, we consider the number of conversions tracked within three days after ad impressions as the actual conversion number.
We also present the conversion delay of paid objectives, please refer to Appendix \ref{sec:delay} for details.
Each sample in Kuai-AD corresponds to the features of an ad campaign and the number of ad conversions. The features can be divided into sparse features and dense features. Sparse features include ad campaign profiles such as optimization objectives, the product name and industry information. Dense features include the number of clicks, the number of views, the number of real-time tracked ad conversions, etc. 
\subsubsection{Metrics}
On the Kuaishou platform, when the advertiser's cost rate is greater than 1.2, the platform will compensate the advertiser; and when the cost rate is less than 0.8, it means that the platform has a large loss in profit. Therefore, our primary goal is to increase the compliance rate (CR),
which is defined as follows:
\begin{equation}
CR = \frac{1}{N}\sum_{y}^{\mathcal{Y}}{\eta(0.8 \leq \frac{\hat{y}}{y} \leq 1.2)}
\end{equation}
where $\eta(\cdot)$ is an indication function, such that $\eta(\text{true}) = 1$ and $\eta(\text{false}) = 0$.
In addition to CR, Mean Absolute Percentage Error (MAPE) is employed to evaluate the percentage error.
\subsubsection{Baselines}
We compared our method with several baseline methods, which are categorized as follows: 
Aggregating pCTCVR of ad impressions: \textbf{RM}; Traditional value regression methods: \textbf{VR}; Parametric distribution regression methods: \textbf{ZILN} \citep{wang2019deep}; Ordinal regression methods: \textbf{CREAD} \citep{sun2024cread}; Bucket classification regression methods: \textbf{MDME} \citep{li2022billion}, \textbf{TPM} \citep{lin2023tree}. 
For our method \textbf{LDACP}, we empirically set $\alpha = 1$, $\beta = 1$, $\psi(y, m)$, $h(\cdot)$ as below:
\begin{equation}
    \psi(y, m) = \frac{|y - m|}{y + \epsilon}
\end{equation}
\begin{equation}
    h(x) = \text{e}^{-10x}
\end{equation}
For a detailed description of the baseline methods, please refer to Appendix \ref{sec:baselines}.
To ensure a fair comparison, all experiments are configured with a batch size of 128 and utilize the Adam optimizer \citep{kingma2014adam} with a learning rate of 1e-3. All networks employ SELU \citep{klambauer2017self} as the activation function, with a consistent parameter initialization method. The maximum number of training epochs is set to 20. An early stop strategy is employed, terminating training if the validation MAPE fails to decrease over two consecutive epochs. The test results from the epoch with the lowest validation MAPE are considered the final experimental results.
Excluding LDACP and RM, we use the number of conversions and PCOC as training labels, denoted as \textbf{method-N} and \textbf{method-P}, respectively.
\subsubsection{Performance Comparison}
As shown in Table \ref{tab:offline_exp}, the results of the offline experiments indicate that:
\begin{enumerate}[leftmargin=*]
    \item Our method outperforms existing competitive methods across various industries in both MAPE and CR. Specifically, the MAPE decreased by 0.0686 compared to the best baseline CREAD-P, and the CR improved by 8.52\% over CREAD-P.
    \item All methods significantly outperform RM, demonstrating the advantages of leveraging post-impression features and conducting regression predictions at the campaign level.
    \item The performance of ZILN-P significantly declines compared to ZILN-N, indicating that ZILN struggles to perform well when the distribution does not conform to its prior assumptions.
    \item The performance of CREAD-N significantly declines compared to CREAD-P because the bucket classification label definition of CREAD is not suitable for scenarios with severe long-tail distributions where the labels are integers.
\end{enumerate}
We also present the experimental results for the CR metric under other thresholds. Please refer to the Appendix \ref{sec:cr}.
\subsection{Ablation Study}
We conducted ablation studies on Kuai-AD to further demonstrate the effectiveness of our label smoothing method and VRMP.
\begin{itemize}[leftmargin=*]
        \item \textbf{-wo-smoothing}: This configuration refers to the LDACP without applying the label smoothing method.
    \item \textbf{-wo-VRMP} means removing the VRMP,  and adopting $\hat{y}_{f}$ as the inference result.
\end{itemize}
\begin{table}[htbp]
    \caption{Ablation study on Kuai-AD}
    \centering
    \begin{tabular}{lccc}
        \toprule
        Metric          & -wo-smoothing   & -wo-VRMP &  LDACP    \\
        \midrule
        MAPE             & 0.2692     & 0.3119      &  \textbf{0.2228}  \\
        CR    & 57.38\%& 52.90\% &  \textbf{62.32\%}  \\
        \bottomrule
    \end{tabular}
    \label{tab:ablation}
\end{table}
Table \ref{tab:ablation} shows that the MAPE for -wo-smoothing increases by 0.0464 compared to LDACP and the CR decreases by 4.94\%, indicating the effectiveness of our label smoothing method. The MAPE for -wo-VRMP increases by 0.0891 compared to LDACP, and the CR decreases by 9.42\%, demonstrating the effectiveness of the VRMP.
\begin{figure*}[ht]
    \centering
    \includegraphics[width=\textwidth]{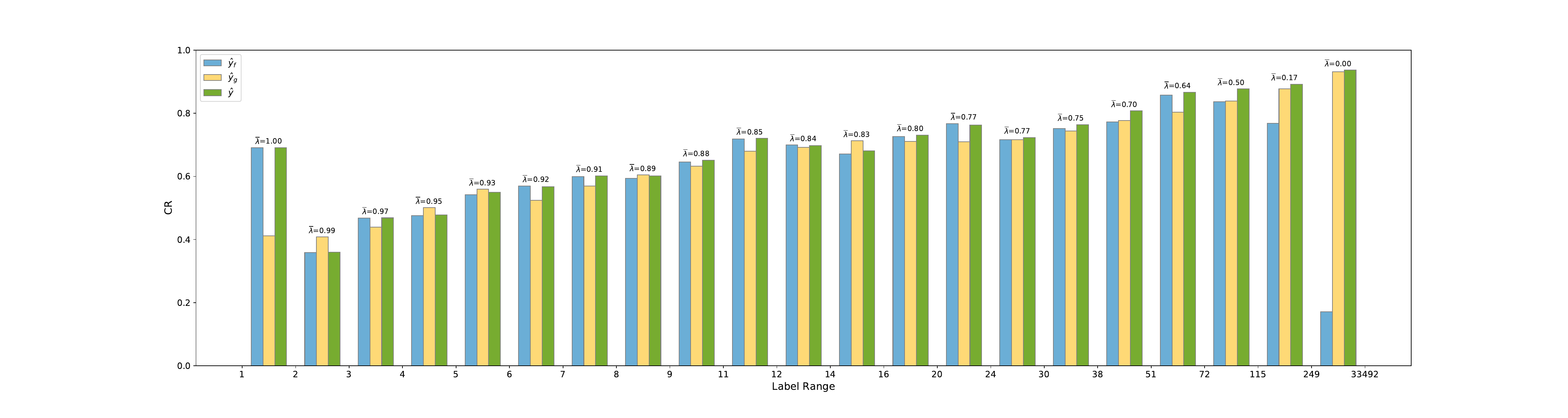}
    \caption{The CR and average value of $\lambda$ on Kuai-AD test set. $\hat{y}_{f}$ and $\hat{y}_{g}$ each have their distinct advantages. $\hat{y}$ integrates the strengths of both to address the challenge of predicting tail data and enhance overall performance.}
    \label{fig:ensemble_moe.pdf}
    \Description{test}
\end{figure*}
We further analyzed the characteristics of $\hat{y}_{f}$, $\hat{y}_{g}$, and $\hat{y}$ in the test set to demonstrate the role of VRMP and MoE in addressing the challenges of long-tail data prediction and improving overall performance. We divided samples of the test set into buckets based on their labels and calculated the CR for each bucket. Figure \ref{fig:ensemble_moe.pdf} shows that 
$\hat{y}_{f}$ has a higher CR for the interval with a smaller label, 
while the $\hat{y}_{g}$ has a higher CR for the interval with a larger label. 
The CR of $\hat{y}$ in most intervals is on par with the best results of $\hat{y}_{f}$ and $\hat{y}_{g}$. We further calculated the average value $\overline{\lambda}$ of the test set samples in each bucket, which reflects the weights of $\hat{y}_{f}$ and $\hat{y}_{g}$ in each bucket interval. It can be seen that $\overline{\lambda}$ gradually decreases as the label value increases, which once again shows that the MoE structure effectively integrates the advantages of both $\hat{y}_{f}$ and $\hat{y}_{g}$, achieving a higher CR overall.
We also explored how the number of leaf nodes affects the performance of LDACP. For more details, please see Appendix \ref{sec:bucket}.
\subsection{Online A/B Test}
\begin{figure}[h]
    \centering
    \includegraphics[width=\linewidth]{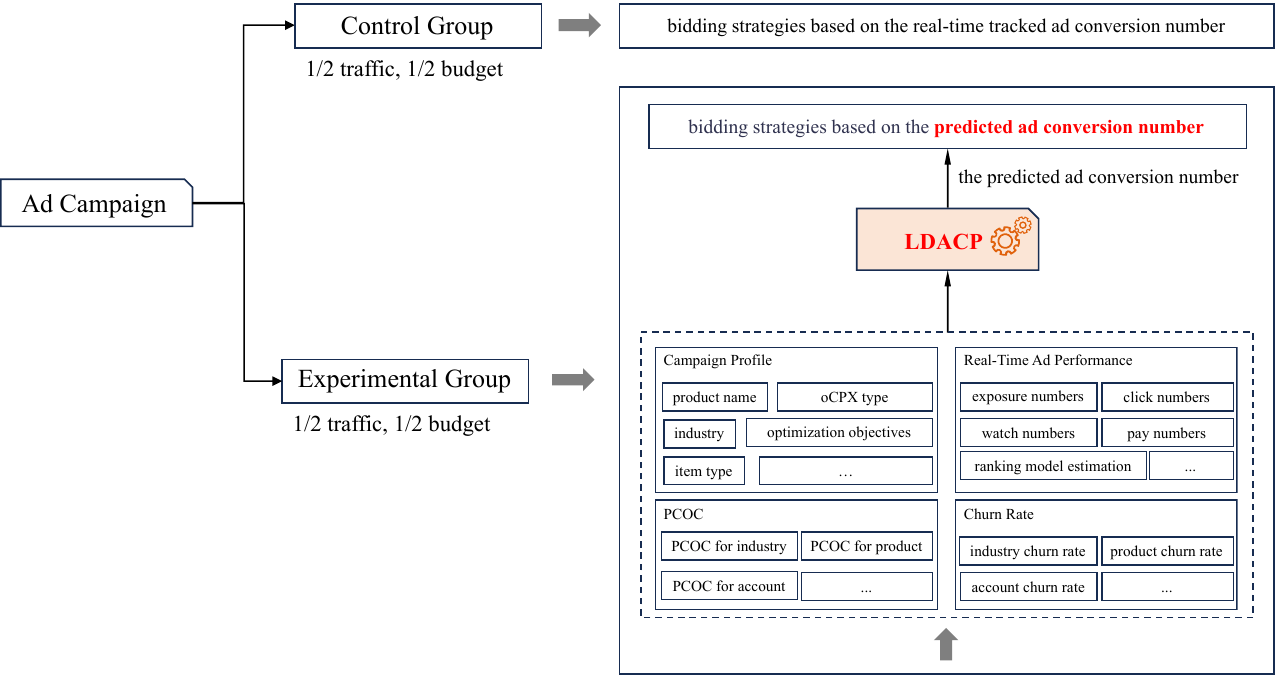}
    \caption{An ad campaign is split into a control group and an experimental group, each receiving half of the traffic and budget. The control group employs bidding strategies based on the number of real-time tracked ad conversions, while the experimental group employs bidding strategies based on the predicted number of conversions.} 
    \label{fig:online_exp}
    \Description{test}
\end{figure}

We conducted an A/B test in the Kuaishou oCPM bidding advertising scenario to evaluate the effectiveness of our method in enhancing advertising performance. Each ad campaign is divided into two groups, each with half of the traffic and budget. One group serves as the control group, employing bidding strategies based on the real-time tracked conversion number; the other group serves as the experimental group, employing bidding strategies based on the predicted conversion number.
Since our offline experiments have demonstrated that LDACP achieves significant performance improvements over the baselines, and considering the substantial commercial impact of conducting online experiments, we will use only LDACP for predicting the number of ad conversions.
As illustrated in Figure \ref{fig:online_exp}, the features of LDACP are categorized into four types. The first category is the campaign profile, such as the product name, optimization objectives and the industry of the ad campaign.
The second category includes features reflecting the effect of advertising, such as the number of impressions, the number of views, the number of clicks and the aggregated pCTCVR of ad impressions. These features undergo a base-10 logarithmic transformation to standardize the feature range before being input into the model.
The third category includes features reflecting the bias of the ranking model, such as the PCOC of the ranking model for industry, product and account.
The fourth category includes features reflecting the churn rate, such as the churn rate of industry, product and account.
For sparse features, we set the embedding dimension to 8.
We ensure the predicted ad conversion number of LDACP is not less than the number of real-time tracked conversions.
During the five-day online A/B experiment, as shown in Table \ref{tab:online_exp_performance}, the experimental group consistently achieved a higher CR than the control group, with an overall improvement of 2.29\%, demonstrating the effectiveness of our method in enhancing the compliance rate.
Additionally, we observed that the revenue of the platform in the experimental group increased by 11.06\% compared to the control group, and the number of ad conversions increased by 8.68\%. 
We analyzed how improving the CR can boost both the platform's revenue and the number of conversions; for more details, refer to Appendix \ref{sec:analysis}.
\begin{table}[h]
    \caption{Cost compliance rate of the online A/B test. The experimental group consistently achieved a higher compliance rate compared to the control group.}
    \centering
    \begin{tabular}{lcc}
        \toprule
        Date              & Control Group          & Experimental Group    \\
        \midrule
        D1        & 53.60\%                       & 58.26\% (4.66\% $\uparrow$)  \\
        D2        & 58.18\%                       & 59.80\% (1.62\% $\uparrow$)  \\
        D3        & 59.85\%                       & 60.45\% (0.59\% $\uparrow$)  \\
        D4        & 58.91\%                       & 59.84\% (0.93\% $\uparrow$)  \\
        D5        & 56.64\%                       & 60.12\% (3.48\% $\uparrow$)  \\
        Total             & 57.38\%               & 59.66\% (2.29\% $\uparrow$)  \\
        \bottomrule
    \end{tabular}
    \label{tab:online_exp_performance}
\end{table}
\section{Conclusion}
In this paper, we propose a novel \textbf{L}ong-\textbf{D}elayed \textbf{A}d \textbf{C}onversions \textbf{P}rediction model for bidding strategy (LDACP) designed to predict the conversion number of an ad campaign and thus determine the bidding strategy of the automated bidding system. LDACP consists of a \textbf{B}ucket \textbf{C}lassification \textbf{M}odule with label \textbf{S}moothing method (BCMS) and a \textbf{V}alue \textbf{R}egression \textbf{M}odule with \textbf{P}roxy labels (VRMP). The BCMS converts one-hot hard labels into soft labels and fits these soft labels by minimizing classification loss and regression loss, effectively addressing the discontinuity issue that arises when transforming regression problems into multi-class classification problems. The VRMP uses the prediction bias of the aggregated pCTCVR as proxy labels, successfully tackling the challenge of predicting tail data. Then we integrate the strengths of the BCMS and VRMP through a MoE structure to produce the final predicted ad conversion number.
 Offline experiments and the online A/B test conducted on the Kuaishou platform have demonstrated the effectiveness of our method in predicting ad conversion numbers and enhancing advertising performance.
 
\bibliographystyle{unsrtnat}
\bibliography{main.bbl}
\appendix
\section{Conversion Delay for Paid Objectives}
\label{sec:delay}
As shown in Table \ref{tab:delay}, we present the average delay of paid objectives on the Kuaishou platform, showing the average delay from impression to conversion. It is evident that most types of paid ad campaigns experience significant conversion delays. Except for the SITE\_PAGE type, the "p50" conversion delays of other types are 20 minutes or longer. The automated bidding system adjusts the CPA bid approximately every 10 minutes. Therefore, if the real-time tracked conversion number is directly used as a bidding strategy adjustment signal, it will lead to a more conservative bidding strategy, making it difficult for ad campaigns to obtain sufficient traffic, which is particularly obvious in the early stages of ad campaign deployment.
\begin{table}[h]
\caption{Conversion delay for paid objectives, where "p10" denotes the time delay for the fastest 10\% of conversions, and other metrics follow the same pattern. The time units in the table are given in minutes.} 
\label{tab:delay}
\centering  
\footnotesize
\begin{tabular}{cccccccccc}  
\toprule  
Campaign Type                   & p10   & p20 & p30 & p40 & p50 & p60 & p70 & p80 & p90 \\
\midrule   
APP                             &  7	& 11  & 17	& 27  &  46	& 103 & 305	& 999 & 2770 \\
APP\_ADVANCE                    &  1	&  3  & 5   & 7   &  20	& 88  & 294 & 569 & 931  \\
SITE\_PAGE                      &  3	&  4  & 5	& 7	  &   9	& 13  & 18	& 29  & 74  \\
LIVE\_STREAM	                &  9    &  18 & 26  & 35  & 50  &  85 & 269 &1176 & 4434 \\
\bottomrule
\end{tabular}
\end{table}
\section{More Details of Baselines}
\label{sec:baselines}
\begin{itemize}[leftmargin=*]
    \item \textbf{RM(Ranking Model)} aggregates the pCTCVR of ad impressions in an ad campaign as the predicted ad conversion number.
    \item \textbf{VR(Value Regression)} directly predicts the number of ad conversions by employing Mean Absolute Error (MAE) as the loss function.
    
    \item \textbf{ZILN} \citep{wang2019deep} models the mean and variance of a log-normal distribution through neural networks, with the expectation of the log-normal distribution serving as the predicted value.
    We adopt the open-source code and parameter configurations provided in the original paper.
    \item \textbf{CREAD} \citep{sun2024cread} predicts video watch time using a multi-class classification approach and introduces the Error-Adaptive Discretization Module (EADM) to balance learning and restoration losses. We implemented CREAD based on the original paper's descriptions and hyperparameters. For the EADM module, we selected the smoothing function specified in the paper. To ensure fair comparison, the number of buckets was set to 64.
    
    \item \textbf{MDME} \citep{li2022billion} addresses complex data distributions using a two-stage bucket classification approach. First, the overall data distribution is divided into $K$ sub-distributions, which are subsequently partitioned into $N$ buckets each. Since the original paper did not provide the source code, we implemented the method based on the descriptions outlined in the study. For fair comparison with other methods, we followed the experimental setup in the paper, setting the number of sub-distributions to 2 and evenly splitting each into 32 buckets.
    
    \item \textbf{TPM} \citep{lin2023tree} employs a progressive tree-based approach to derive inference results. Our experiments were conducted using the implementation provided in the original paper. We set the tree height to 7 thus leaf nodes contain 64 buckets. Given the pronounced long-tail distribution of Kuai-AD, using the midpoint of intervals as the expectation value for leaf nodes would lead to significant overestimation. Therefore, we calculate the expectation values of the leaf nodes following the method outlined in \citep{fu2018deep}.
    The TPM paper introduces node loss, prediction loss and variance loss terms. However, due to the wide range of ad conversion numbers, using both node loss and variance loss terms would hinder model training. Therefore, we only employed the node loss term.
\end{itemize}
\section{Cost Compliance Rate Under Different Thresholds}
\label{sec:cr}
\begin{figure}[htbp]
    \centering
    \includegraphics[width=\linewidth]{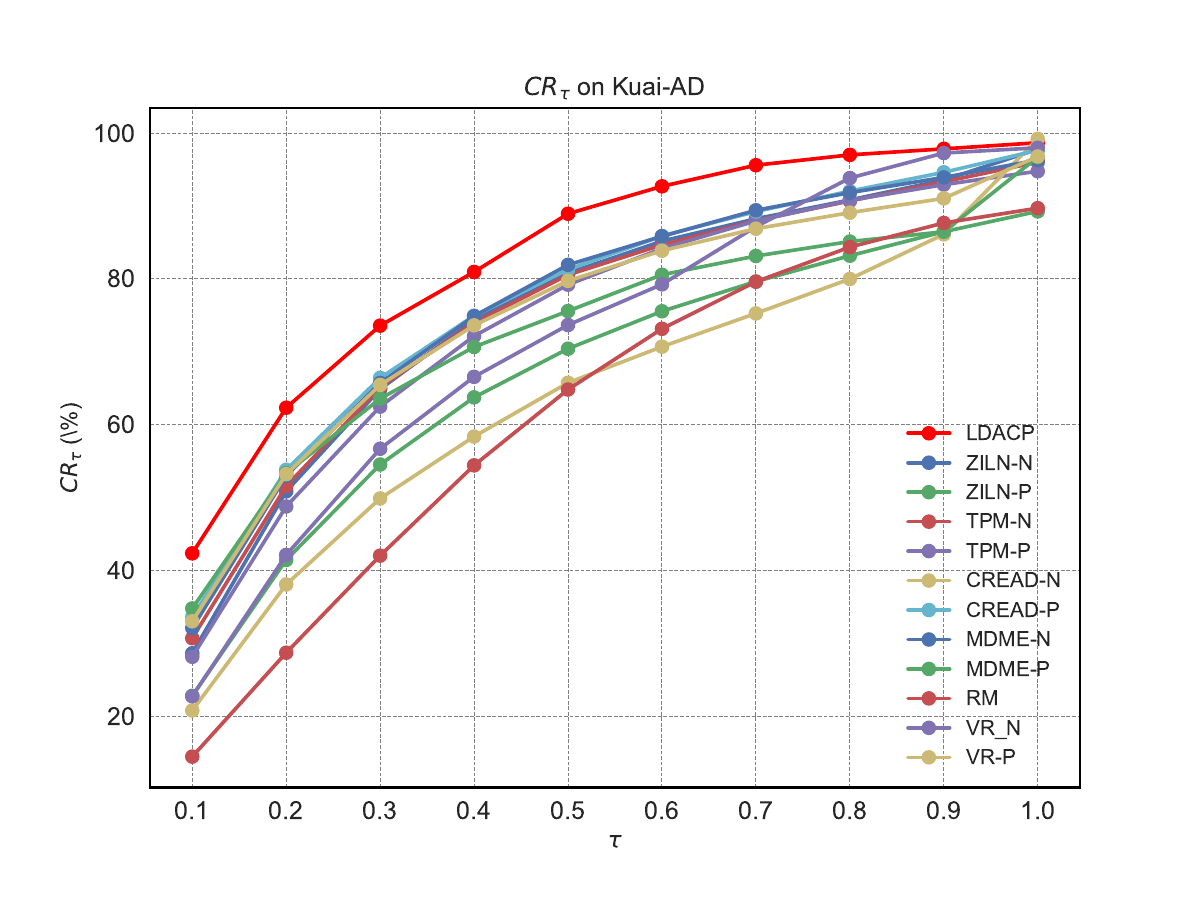}
    \caption{Comparison of the $CR_{\tau}$ between LDACP and baseline methods on Kuai-AD test set.} 
    \label{fig:CR_plot}
    \Description{test}
\end{figure}
We analyzed the cost compliance rate of LDACP and baseline methods under varying thresholds. This experiment was conducted on Kuai-AD dataset, using the same experimental configuration as described in Section \ref{sec:experiment}.
We define $CR_{\tau}$ as:
\begin{equation}
    CR_{\tau} = \frac{1}{N}\sum_{y}^{\mathcal{Y}}{\eta(1 - \tau \leq \frac{\hat{y}}{y} \leq 1 + \tau)}
\end{equation}
As illustrated in Figure \ref{fig:CR_plot}, LDACP consistently outperforms other baseline methods across a range of threshold values for $\tau$, demonstrating its robustness and effectiveness.
\section{The Impact of the Number of Leaf Nodes on the Performance of LDACP}
\label{sec:bucket}
\begin{figure}[htbp]
    \centering
    \begin{subfigure}[b]{\linewidth}
        \centering
        \includegraphics[width=0.8\linewidth]{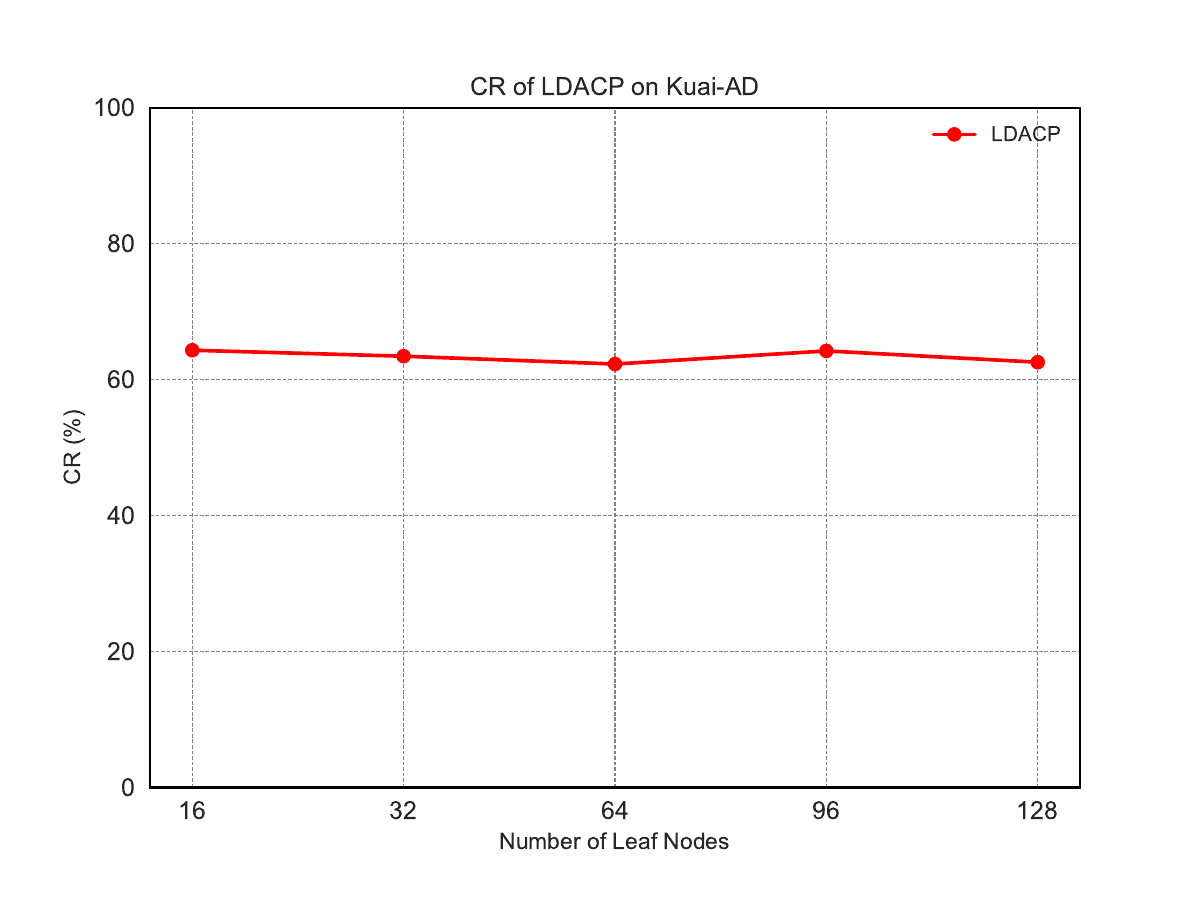}
        \caption{CR of LDACP with varying numbers of leaf nodes.}
        \label{fig:bucket_cr}
    \end{subfigure}
    \begin{subfigure}[b]{\linewidth}
        \centering
        \includegraphics[width=0.8\linewidth]{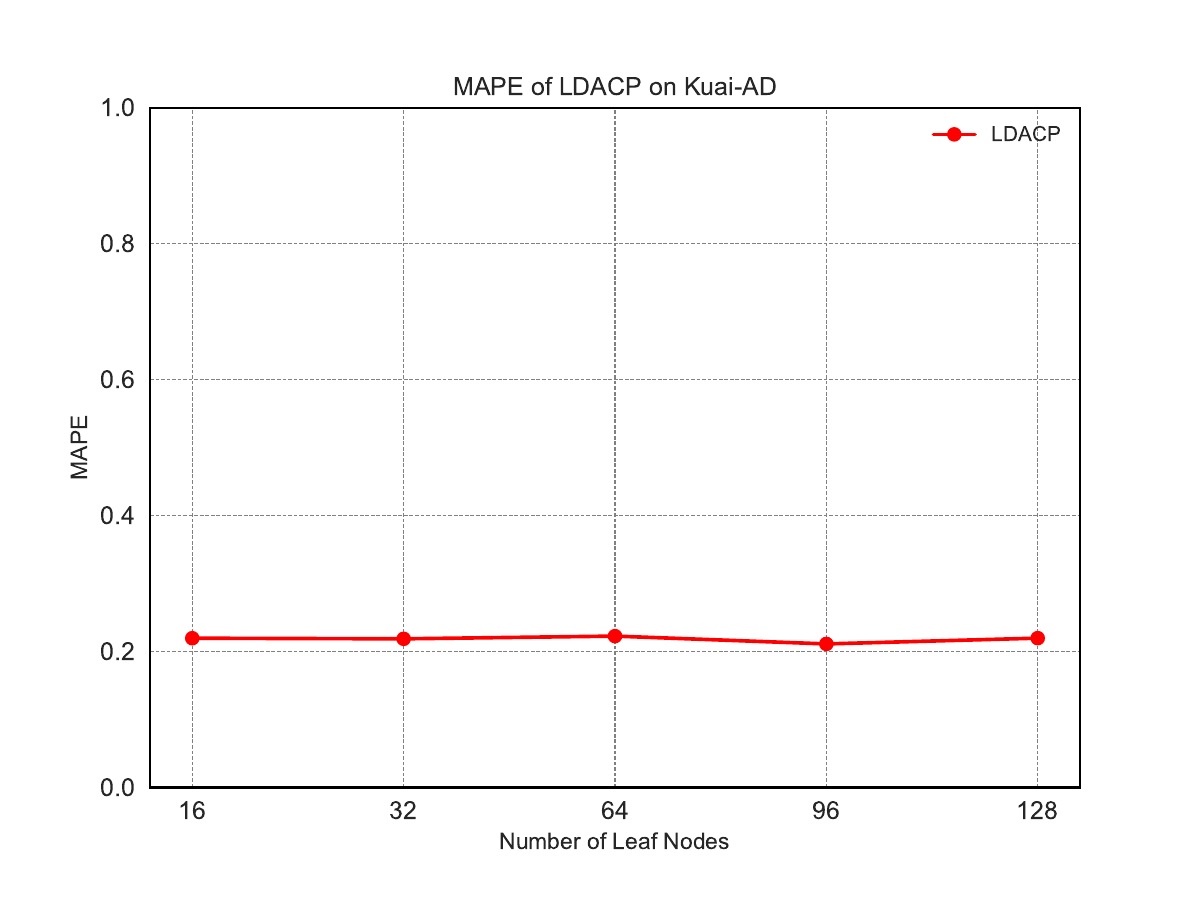}
        \caption{MAPE of LDACP with varying numbers of leaf nodes.}
        \label{fig:bucket_mape}
    \end{subfigure}
    \caption{Performance of LDACP with varying numbers of leaf nodes on Kuai-AD.}
    \label{fig:bucket_exp_res}
    \Description{test}
\end{figure}

We studied the performance of LDACP on Kuai-AD with different numbers of leaf nodes. Specifically, we set the number of leaf nodes of TPMS to \{16, 32, 64, 96, 128\}. As illustrated in Figure \ref{fig:bucket_exp_res}, the change in the number of leaf nodes has a certain impact on the CR and MAPE metrics, but the impact is relatively small.
\section{Analysis of Online Ad Performance}
\label{sec:analysis}
The online experiment showed increases in cost compliance rate, platform revenue, and ad conversions.
Here, we explain how improving the cost compliance rate can increase advertising platform revenue and boost ad conversions.
Since the number of real-time tracked ad conversions is not greater than the actual number of ad conversions, the current CPA will be overestimated.
To prevent the current CPA from exceeding the target CPA, the automated bidding system adopts a more conservative strategy.
In severe cases, this may cause ad campaigns to lose in traffic auctions and become inactive.
Inactive ad campaigns generate no revenue for the platform, and advertisers fail to meet their target advertising effects.
Our method addresses this issue by accurately predicting the number of ad conversions, enabling the automated bidding system to use these predictions as decision-making signals.
This enables the development of more effective bidding strategies, enhancing ad campaign performance. 
Consequently, the platform gains more profit, and advertisers achieve their target advertising effects.
\end{document}